\pdfoutput=1
\documentclass[sigconf,screen]{acmart}

\AtBeginDocument{%
  }
\usepackage{multirow}
\usepackage{pifont} 
\usepackage{subcaption}
\usepackage{xcolor}
\usepackage{paracol}

\acmConference[MM '25]{October 27--31,
  2025}{Dublin, Ireland}

\begin{document}

\title{LDGNet: A Lightweight Difference Guiding Network for Remote Sensing Change Detection}

\settopmatter{printacmref=false}

\author{Chenfeng Xu}
\affiliation{%
  \institution{China University of Mining and Technology}
  \city{Xuzhou}
  \state{Jiangsu}
  \country{China}}

\begin{abstract}
With the rapid advancement of deep learning, the field of change detection (CD) in remote sensing imagery has achieved remarkable progress. Existing change detection methods primarily focus on achieving higher accuracy with increased computational costs and parameter sizes, leaving development of lightweight methods for rapid real-world processing an underexplored challenge. To address this challenge, we propose a Lightweight Difference Guiding Network (LDGNet), leveraging absolute difference image to guide optical remote sensing change detection. First, to enhance the feature representation capability of the lightweight backbone network, we propose the Difference Guiding Module (DGM), which leverages multi-scale features extracted from the absolute difference image to progressively influence the original image encoder at each layer, thereby reinforcing feature extraction. Second, we propose the Difference-Aware Dynamic Fusion (DADF) module with Visual State Space Model (VSSM) for lightweight long-range dependency modeling. The module first uses feature absolute differences to guide VSSM's global contextual modeling of change regions, then employs difference attention to dynamically fuse these long-range features with feature differences, enhancing change semantics while suppressing noise and background. Extensive experiments on multiple datasets demonstrate that our method achieves comparable or superior performance to current state-of-the-art (SOTA) methods requiring several times more computation, while maintaining only 3.43M parameters  and 1.12G FLOPs. Code will be released after revision.

\end{abstract}

\begin{CCSXML}
<ccs2012>
 <concept>
  <concept_id>00000000.0000000.0000000</concept_id>
  <concept_desc>Do Not Use This Code, Generate the Correct Terms for Your Paper</concept_desc>
  <concept_significance>500</concept_significance>
 </concept>
 <concept>
  <concept_id>00000000.00000000.00000000</concept_id>
  <concept_desc>Do Not Use This Code, Generate the Correct Terms for Your Paper</concept_desc>
  <concept_significance>300</concept_significance>
 </concept>
 <concept>
  <concept_id>00000000.00000000.00000000</concept_id>
  <concept_desc>Do Not Use This Code, Generate the Correct Terms for Your Paper</concept_desc>
  <concept_significance>100</concept_significance>
 </concept>
 <concept>
  <concept_id>00000000.00000000.00000000</concept_id>
  <concept_desc>Do Not Use This Code, Generate the Correct Terms for Your Paper</concept_desc>
  <concept_significance>100</concept_significance>
 </concept>
</ccs2012>
\end{CCSXML}

\ccsdesc[500]{Computing methodologies~Computer vision}

\keywords{ Change detection (CD), remote sensing (RS), state space model, difference-guidance, lightweight architecture}

\begin{teaserfigure}
    \centering
    \begin{subfigure}[b]{0.33\textwidth}
        \includegraphics[width=\textwidth]{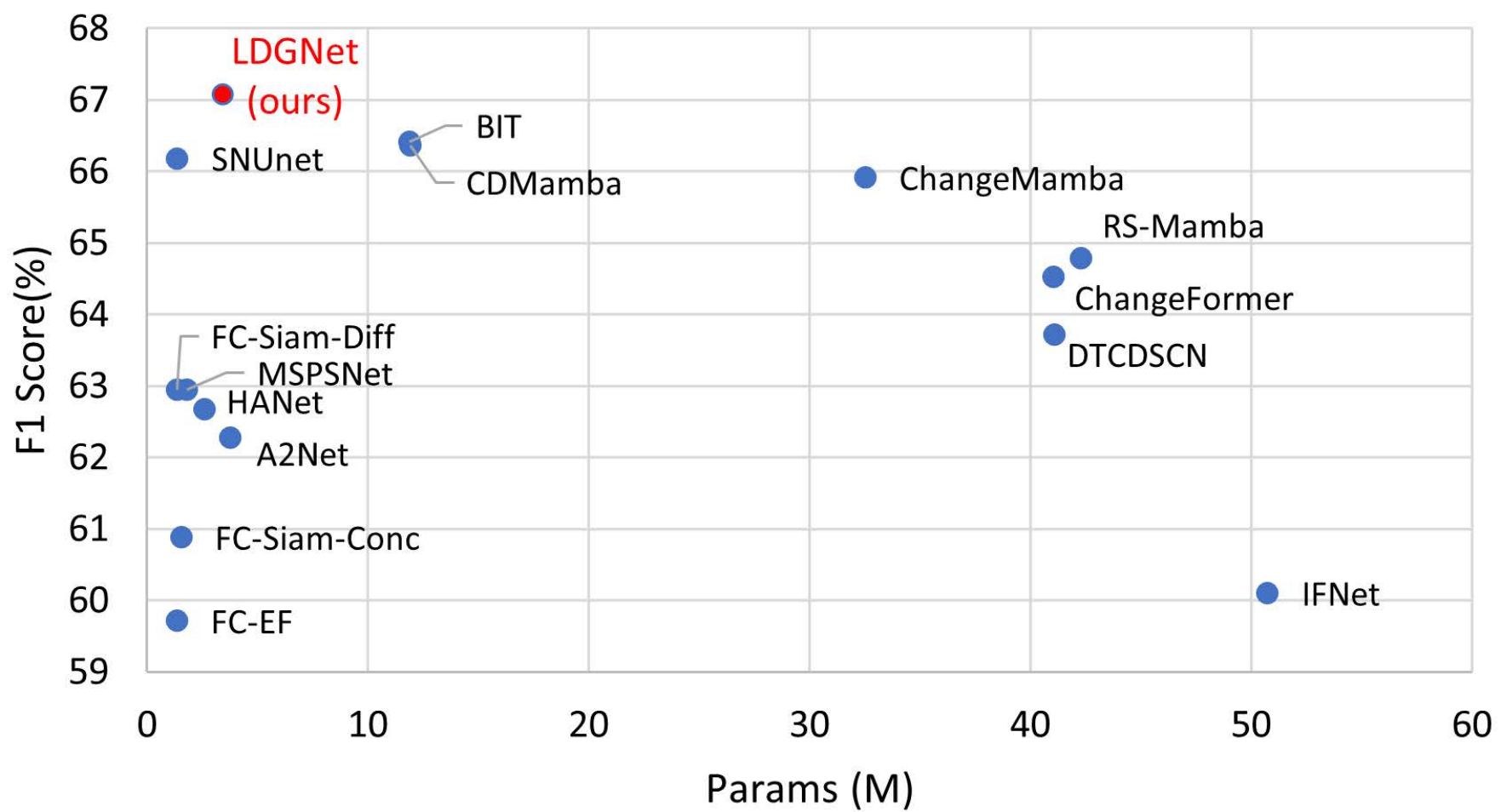}
        \caption{Params Comparison}
        \label{fig:params}
    \end{subfigure}
    \hfill
    \begin{subfigure}[b]{0.33\textwidth}
        \includegraphics[width=\textwidth]{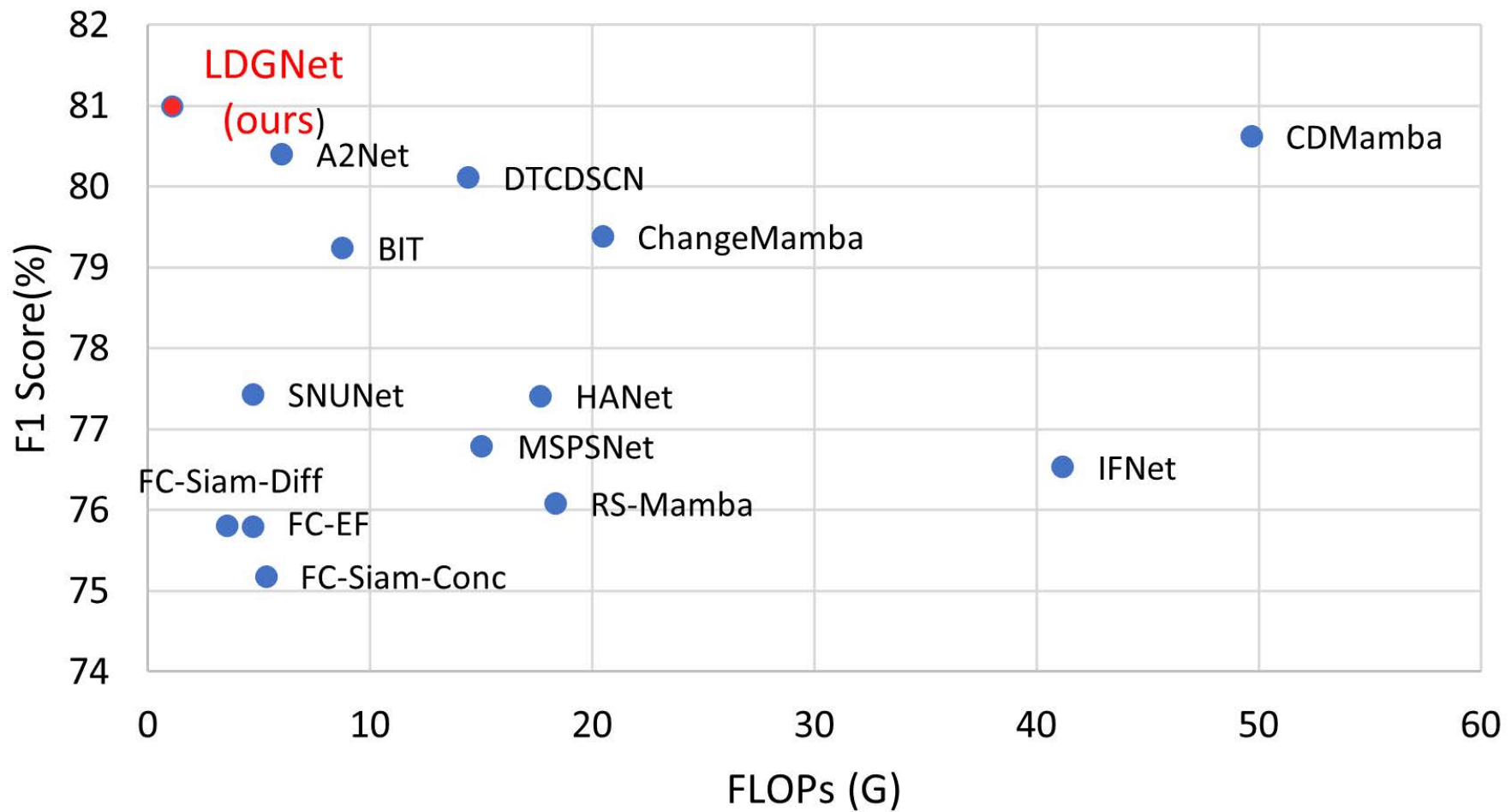}
        \caption{FLOPs Comparison}
        \label{fig:flops}
    \end{subfigure}
    \hfill
    \begin{subfigure}[b]{0.33\textwidth}
        \includegraphics[width=\textwidth]{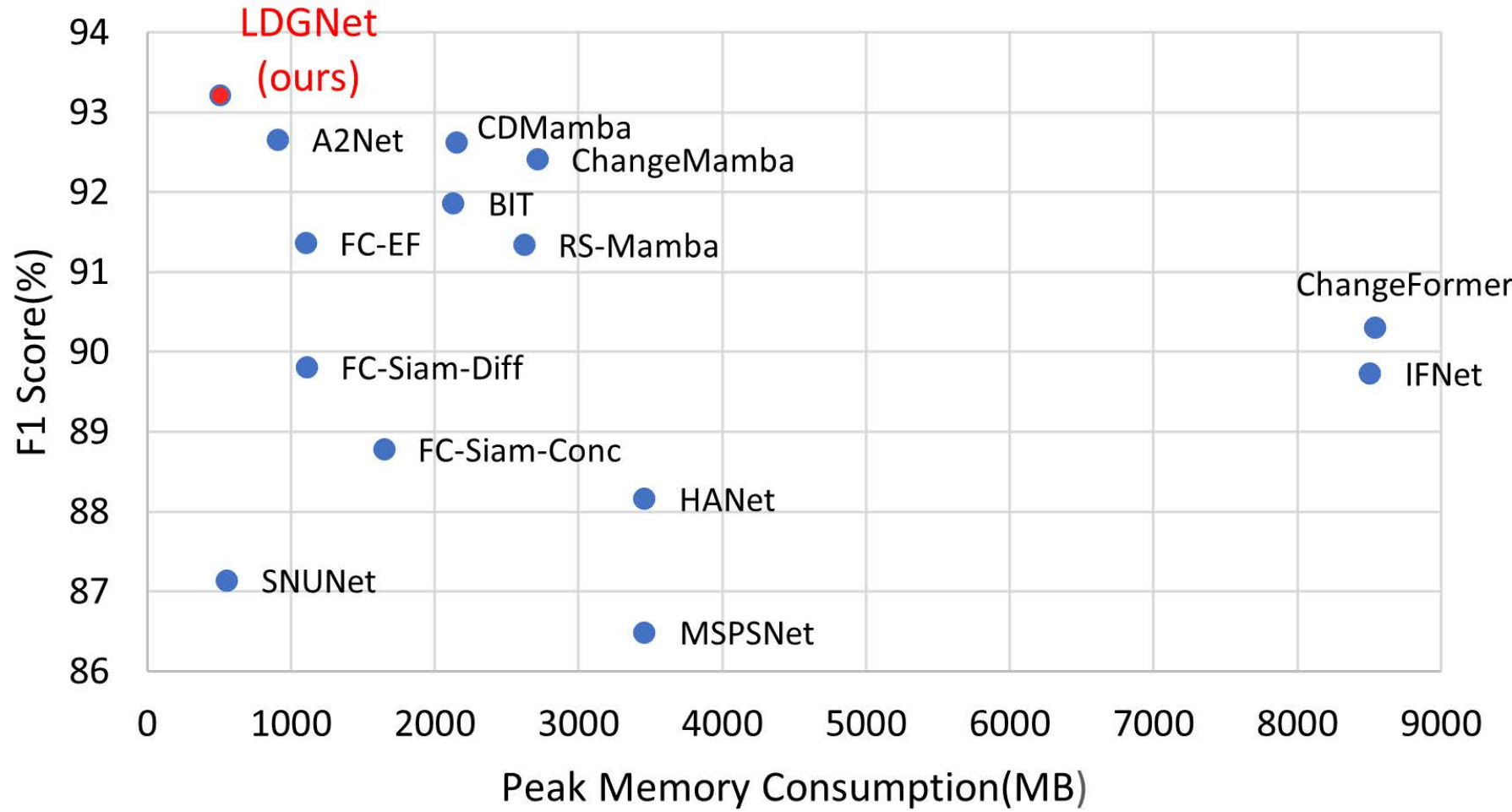}
        \caption{Memory Comparison}
        \label{fig:memory}
    \end{subfigure}
    \caption{Comparative analysis of model parameters, FLOPs and memory consumption. The Y-axis (from left to right) sequentially represents the model's performance on the DSIFN-CD dataset, SYSU-CD dataset, and WHU-CD dataset, respectively.}
    \label{fig:comparison}
\end{teaserfigure}

\maketitle

\section{Introduction}
Change detection in remote sensing images is designed to predict regions of alteration by comparing co-registered images captured at different time points \cite{lu2004change}. In recent years, with the rapid development of deep learning, significant advancements have been achieved in the field of change detection \cite{coppinp2004digital}. Initially, change detection methods based on Convolutional Neural Networks (CNNs) \cite{daudt2018fully,ifnet,dtcdscn,fang2021snunet,han2023hanet,MSPS,a2,lcd,rfa} were widely adopted due to their excellent ability to capture local details. However, the inherently limited receptive field of CNNs makes it difficult to capture global semantic relationships in remote sensing imagery \cite{chen2024changemamba}. 
Subsequently, Transformer-based methods \cite{BIT,cf,swinsunet,li2022transunetcd}, leveraging self-attention mechanisms \cite{vaswani2017attention}, have demonstrated superior capability in extracting global semantic dependencies, achieving better detection performance than CNNs. However, the quadratic computational complexity of Transformers leads to prohibitively high computational costs, making them difficult to apply to large-scale remote sensing data. To mitigate this, Transformer-based change detection methods often restrict window sizes to control computational demands, but this inevitably limits the model's receptive field \cite{chen2024changemamba}.
Reviewing the evolution of deep learning-based change detection methods, achieving a balance between detection accuracy and computational efficiency remains a persistent research challenge in the field \cite{mmdetection}.

State space models \cite{gu2021efficiently} have demonstrated remarkable advantages in processing sequential data. The Mamba architecture \cite{gu2023mamba}, based on this theory, employs a selective scanning mechanism to achieve global modeling of long sequence data while maintaining linear computational complexity. Notably, similar to Transformer, visual variants of the Mamba architecture (e.g. Vision Mamba \cite{zhu2024visionmamba} and VMamba \cite{liu2024vmamba}) have also been applied to change detection in remote sensing imagery. The inherent cumulative mechanism of state-space models, provides a solid theoretical foundation for simultaneously balancing local feature extraction and global semantic representation.
Although Mamba-based change detection methods \cite{chen2024changemamba,rsmamba,cdmamba} have achieved remarkable performance while maintaining linear computational complexity and have attracted extensive attention in recent years, their computational requirements remain high compared to truly lightweight architectures \cite{lcd,rfa}. Currently, most existing detection methods still rely on increasingly complex network structures to obtain more precise measurement results. How to achieve relatively good detection performance under limited computational resources to meet the demands of practical applications remains an underexplored research topic.

Current lightweight detection methods remain predominantly governed by CNNs. We summarize two limitations of existing lightweight methods: (1) Use of lightweight backbone networks inevitably introduces the problem of limited feature representation capability. Information lost during the feature extraction stage is inherently difficult to recover in subsequent processing phases \cite{rfa}. Many existing approaches typically attempt to alleviate this issue during the feature fusion stage by employing strategies such as multi-level feature aggregation \cite{a2}, while rarely exploring ways to enhance the intrinsic feature extraction ability of the backbone network itself. (2) Given the quadratic computational complexity of global relationship modeling in Transformer architectures \cite{vaswani2017attention}, existing methods primarily employ computationally efficient operations such as large-kernel convolutions \cite{li2024lsknet} or dilated convolutions \cite{a2} for long-range dependency modeling. However, their receptive fields remain fundamentally constrained, while simultaneously performing indiscriminate feature modeling across entire feature maps, thereby inefficiently allocating precious computational resources \cite{mmdetection}. 

It is noteworthy that the feature difference method, used in early change detection, explicitly localizes change regions and quantifies their intensity by directly computing pixel-level difference maps \cite{review}. The physical interpretability of this approach is rooted in the inductive bias that "significant changes induce significant pixel differences". In this paper, we reexamine the role of absolute difference images in lightweight change detection. Our intuition is that the incorporation of absolute difference images into the change detection paradigm fundamentally addresses the two limitations above. To address limitation (1), we argue that the feature difference method can serve as a prior-based attention mechanism to compensate for the insufficient feature representation capability of lightweight backbone networks, guiding the model to learn image differences in a more targeted manner, thereby adaptively shifting the model’s focus regions and ultimately enhancing the representational capability of the lightweight model. For limitation (2), the absolute difference between extracted features directs the limited computational resources to regions with higher change probabilities, achieving more efficient and targeted global context modeling with emerging Mamba architecture. Meanwhile, the complementary fusion of concatenated features and difference features enables effective suppression of background noise and pseudo-changes. 

This paper proposes a lightweight difference guiding network  (LDGNet), which aims to achieve high-precision change detection through optimized computational resource allocation. The framework consists of a lightweight encoder-decoder structure: the encoder employs MobileNetV3 \cite{mobilenetv3} as the backbone network. To address its limited representational capacity, we introduce a dedicated encoder branch incorporating absolute difference images, established a hierarchical interaction mechanism between difference image features and original bi-temporal image features. Our method utilizes a Difference Guidance Module (DGM) to implement progressive spatial channel attention guidance for the original bi-temporal image encoding process, effectively enhancing the model's feature representation capability and sensitivity for change regions. The decoder incorporates a Difference-aware Dynamic Fusion (DADF) module, whose core component is the Visual State Space Model (VSSM) \cite{liu2024vmamba} that concurrently modeling both local details and global semantics through its state space formulation with linear complexity. We use feature differences to guide the VSSM for more targeted feature fusion, and then perform secondary fusion through a difference-aware attention mechanism with the original feature differences, enhancing focus on change regions while suppressing background and noise. As shown in \hyperref[fig:comparison]{Fig. 1}, experimental results demonstrate that this difference-guided encoding-decoding architecture achieves a better balance in efficiency and accuracy under the constraints of 3.43M parameters and 1.12G FLOPs computational load.

Our contributions are summarized as follows:

$\bullet$ We introduce a multi-scale difference feature extraction branch and integrated Difference Guidance Module (DGM) to establish a hierarchical feature guidance mechanism to enhance the sensitivity of the lightweight backbone network to regions with higher change probabilities, thereby improving the feature representation capability of the lightweight encoder.

$\bullet$ We propose a Difference-aware Dynamic Fusion Module (DADF), introducing the Visual State Space Model (VSSM) into optical lightweight change detection. The DADF module implements dual mechanisms for difference region focus and noise suppression, combined with dynamic fusion of concatenated features and differential features, achieving efficient decoding that preserves both local details and global semantic correlations.

$\bullet$ Under resource constraints of 3.43M parameters and 1.12G FLOPs, without any complex training strategies, the proposed network architecture achieves detection accuracy comparable to state-of-the-art models requiring several times more computational resources, providing a new technical paradigm for edge computing deployment.

\begin{figure*}
  \includegraphics[width=\textwidth]{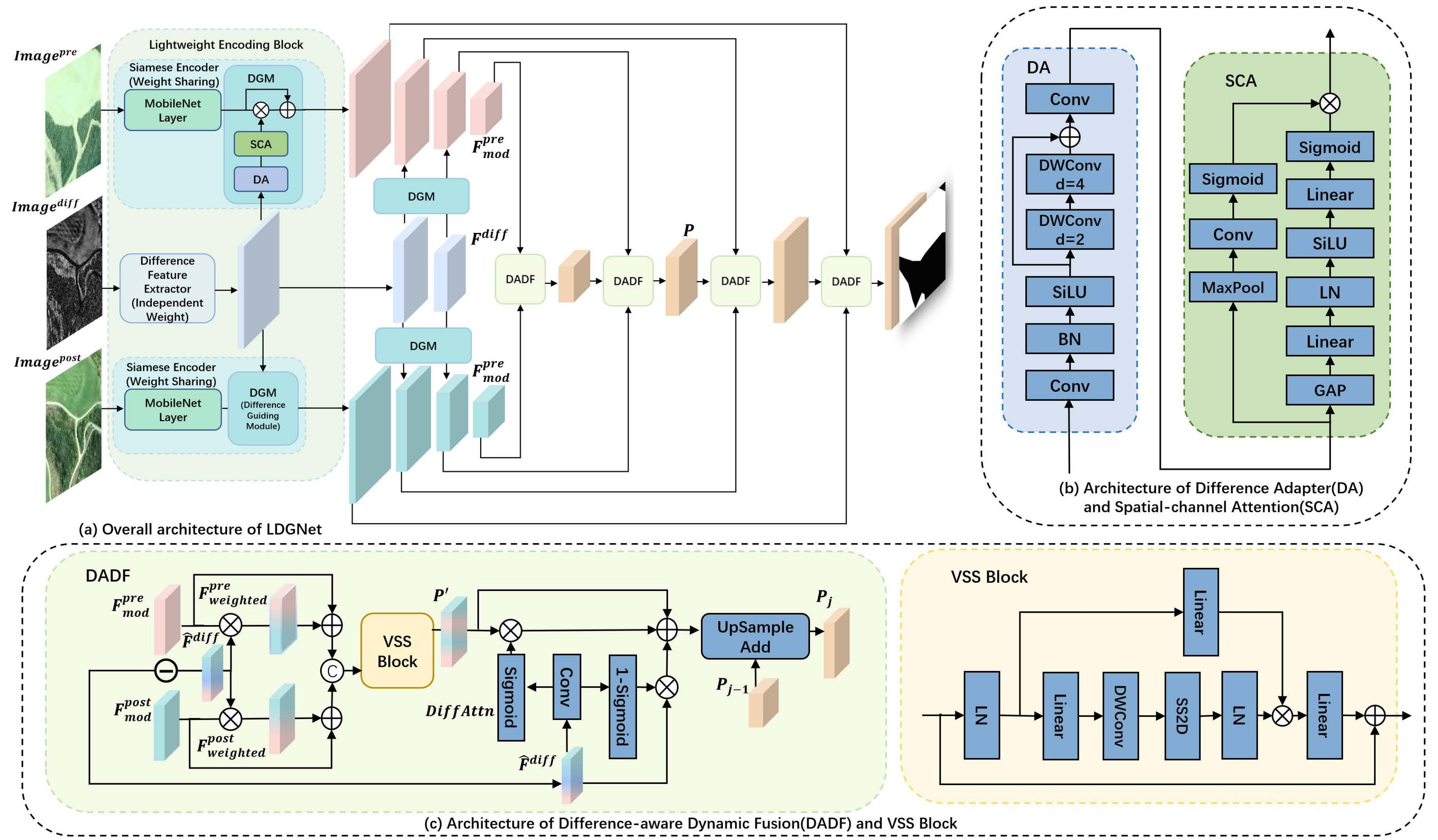}
  \caption{Illustration of our method.
(a) represents overall architecture of our network. The encoder consists of an independent difference feature extractor and a Siamese raw image encoder with shared weights. For each layer, the features extracted independently by the difference extractor are applied to the raw image encoder via the DGM. The decoder is composed of four DADF modules. The pre-event and post-event features are weighted with feature differences and fused through the VSS Block, then dynamically combined with feature differences through DFM. After upsampling and summation with the features from the previous layer, the result is fed into the next DADF module.
(b) represents the detailed structure of the two components of the DGM: DA and SCA.
(c) represents the structure of the DADF and the detailed construction of VSS Block.}

  \label{fig2}
\end{figure*}

\section{RELATED WORK}

\subsection{Deep Learning Methods in CD}
Convolutional Neural Networks (CNNs), owing to their powerful feature extraction capabilities, became the dominant approach in the early stages of remote sensing change detection. Classic architectures include Siamese networks \cite{chicco2021siamese}, U-Net \cite{lv2022simple_unet}, and Fully Convolutional Networks (FCNs) \cite{zhou2023fcn_unet}. However, CNNs are inherently constrained by their fixed receptive fields, making it challenging to effectively capture global semantic information \cite{review}.
The application of Transformers in computer vision has addressed this limitation to some extent. For instance, Chen et al. \cite{BIT} proposed the Bi-temporal Image Transformer (BIT), which models spatiotemporal dependencies by treating images as semantic tokens. Zhang et al. \cite{swinsunet} further designed a purely Transformer-based change detection framework, showcasing the potential of Transformers in change detection tasks. However, the quadratic computational complexity induced by the self-attention mechanism \cite{vaswani2017attention} remains a major bottleneck, making it difficult to apply Transformers to large-scale remote sensing images.

Recently, Mamba-based change detection methods have gained increasing attention due to their ability to maintain global semantic understanding while achieving linear computational complexity \cite{gu2023mamba}. For example, ChangeMamba \cite{chen2024changemamba} utilizes VMamba \cite{liu2024vmamba}, a Mamba variant for vision, for bi-temporal image encoding and decoding, while CDMamba \cite{cdmamba} integrates another Mamba varient, Vision Mamba \cite{zhu2024visionmamba}, with CNNs to balance global information extraction with enhanced local feature modeling. RS-Mamba \cite{rsmamba} introduces omnidirectional scanning to optimize feature extraction for remote sensing imagery. We can see the co-optimization of detection accuracy and computational efficiency has become a focus \cite{mmdetection}. However, in edge computing deployment scenarios, the computational complexity of existing high-precision models still generally exceeds the resource capacity thresholds of edge devices \cite{rfa}. This study pioneers the integration of Visual State Space Models (VSSM) \cite{liu2024vmamba} with lightweight CNNs for optical image change detection. The proposed framework achieves an optimal equilibrium between detection accuracy and computational efficiency through streamlined architectural design, providing a new solution for resource-constrained application scenarios.

\subsection{Feature Difference Guidance in CD}
In traditional change detection (CD) methods, pixel-wise comparison of multi-temporal images is performed to generate difference maps, followed by thresholding to produce change maps \cite{review}. However, this approach heavily relies on empirically determined thresholds. With the rise of deep learning, this method has gradually been replaced by feature concatenation, where models implicitly learn differences. Nevertheless, the traditional difference-based approach inherently provides a more explicit and interpretable representation of change regions, aligning more closely with human perception of change. This suggests its potential as prior knowledge to guide feature extraction and modeling. Research on feature difference guidance in CD has already emerged. For example, TransUNetCD \cite{li2022transunetcd} multiplies the processed difference features with the fused features when generating the final prediction map. However, it does not fully explore the potential guiding role of the feature difference method in both the encoding and decoding stages. CGNet \cite{cgnet} attempts to leverage the semantic information from high-level feature difference maps to guide the fusion of low-level difference maps and still does not explore the potential guiding role that raw image difference maps might play in feature extraction. This study attempts to guide the change detection process from encoding to decoding by utilizing the differences between raw images and extracted features. By developing cross-modal hierarchical interaction and dynamic fusion paradigm between difference features and original bi-temporal features, our methods achieve computational resource efficiency while maintaining detection accuracy comparable to complex models.

\begin{table*}[t]
    \centering
    
    \label{tab:sysu_cd}
    \renewcommand{\arraystretch}{1.1}
    \setlength{\tabcolsep}{2pt}
    \footnotesize
    \caption{Performance comparison between LDGNet and baseline methods on four datasets, with top-performing metrics highlighted in \textcolor{red}{red} and second-best in \textcolor{blue}{blue}.}
    \begin{tabular}{l|ccccc|ccccc|ccccc|ccccc}
        \toprule
         \multirow{2}{*}{\textbf{Method}} 
        
        & \multicolumn{5}{c|}{\textbf{SYSU-CD}} 
        & \multicolumn{5}{c|}{\textbf{LEVIR-CD}}
        & \multicolumn{5}{c|}{\textbf{WHU-CD}} 
        & \multicolumn{5}{c}{\textbf{DSIFN-CD}}\\

        & \textbf{Rec} & \textbf{Pre} & \textbf{OA} & \textbf{F1} & \textbf{IoU}  
        & \textbf{Rec} & \textbf{Pre} & \textbf{OA} & \textbf{F1} & \textbf{IoU}
        & \textbf{Rec} & \textbf{Pre} & \textbf{OA} & \textbf{F1} & \textbf{IoU}  
        & \textbf{Rec} & \textbf{Pre} & \textbf{OA} & \textbf{F1} & \textbf{IoU}\\
        \midrule
        FC-EF~ \cite{daudt2018fully}
        & 75.17 & 76.47 & 88.69 & 75.81 & 61.04 
        & 87.23 & 90.64 & 98.89 & 88.90& 80.03
        & 90.64 & 92.10 & 99.32 & 91.36& 84.10
        & 57.75 & 61.80 & 86.77 & 59.71& 42.56\\
        FC-Siam-Diff~ \cite{daudt2018fully} 
        & 75.30 & 76.28 & 88.65 & 75.79 & 61.01
        & 88.59 & 90.81 & 98.96 & 89.69& 81.31
        & 92.36 & 87.39& 99.16 & 89.81& 81.50
        & 58.27 & 68.44 & 88.35 & 62.95& 45.93\\
        FC-Siam-Conc~ \cite{daudt2018fully} 
        & 76.75 & 73.67 & 88.05 & 75.18 & 60.23
        & 88.43 & 91.41 & 98.98 & 89.89& 81.64
        & 91.11 & 86.57 & 99.08 & 88.78& 79.83
        & 62.80 & 59.08 & 86.30 & 60.88& 43.76\\
        IFNet~ \cite{ifnet}
        & 73.58 & 79.59 & 89.17 & 76.53 & 61.91    
        & 86.65 & 89.62 & 98.81 & 88.11& 78.75
        & 88.01 & 91.51 & 99.20 & 89.73& 81.37
        & 53.94 & 67.86 & 87.83 & 60.10& 42.96\\
        DTCDSCN~  \cite{dtcdscn}
        & 77.25 & 83.19 & 90.96 & 80.11 & 66.82
        & 86.83 & 88.53 & 98.77 & 87.67& 78.05
        & 82.30 & 63.92 & 97.42 & 71.95& 56.19
        & \textcolor{blue}{\textbf{77.99}} & 53.87 & 84.91 & 63.72& 46.76\\
        SNUNet~ \cite{fang2021snunet} 
        & 73.39 & 81.93 & 89.91 & 77.43 & 63.17
        & 87.17 & 89.18 & 98.81 & 88.16& 78.83
        & 89.73 & 84.70 & 98.95 & 87.14& 77.22
        & 72.89 & 60.60 & 87.34& 66.18& 49.45\\
        HANet~ \cite{han2023hanet} 
        & 76.14 & 78.71 & 89.52 & 77.41 & 63.14
        & 89.36 & 91.21 & \textcolor{red}{\textbf{99.02}} & 90.28& 82.27
        & 88.01 & 88.30 & 98.96 & 88.16& 78.82
        & 70.33 & 56.52 & 85.76 & 62.67& 45.64\\
        MSPSNet 
        & 76.29 & 77.29 & 89.06 & 76.79 & 62.32
        & 88.61 & 90.75 & 98.96 & 89.67& 81.27
        & 85.17 & 87.84 & 98.98 & 86.49& 76.19
        & 73.92 & 54.81 & 86.67 & 62.95& 45.93\\
        A2Net~ \cite{a2} 
        & 77.71 & 83.27 & 90.76 & 80.40 & 67.22
        & 88.95 & 91.34 & 98.95 & 90.13& 82.03
        & 91.23 & \textcolor{blue}{94.16} & \textcolor{red}{\textbf{99.51}} & \textcolor{blue}{\textbf{92.66}}& \textcolor{blue}{\textbf{86.33}}
        & 57.32 & 68.17 & 85.34 & 62.28& 45.22\\
        STANet~ \cite{huang2024sei} 
        & 78.10 & 78.61 & 89.92 & 77.80 & 63.71
        & \textcolor{red}{\textbf{90.10}} & 80.81 & 98.54 & 85.20& 74.22
        & 89.30 & 75.70 & 98.36 & 82.00& 69.44
        & 67.71 & 61.68 & 88.49 & 64.50& 47.80\\
        LCD-Net~  \cite{lcd}
        & 77.10 & 84.17 & 90.84 & 80.48 & 67.33
        & 85.98 & \textcolor{red}{\textbf{92.64}} & 98.90 & 89.18& 80.48
        & 92.09 & 92.09 & 99.37 & 92.09& 85.34
        & 54.14 & \textcolor{red}{\textbf{77.86}} & 86.50 & 63.87& 46.91\\
        RFANet~  \cite{rfa}
        & \textcolor{blue}{\textbf{78.21}} & 82.65 & \textcolor{red}{\textbf{91.77}} & 80.37 & 67.18
        & 88.55 & \textcolor{blue}{\textbf{92.12}} & 99.00 & \textcolor{blue}{\textbf{90.33}}& \textcolor{blue}{\textbf{82.37}}
        & 92.40 & 91.24 & 99.35 & 91.81& 84.87
        & 63.92 & 69.65 & 87.90& \textcolor{blue}{\textbf{66.66}}& \textcolor{blue}{\textbf{49.99}}\\
        BIT~ \cite{BIT} 
        & 74.29 & \textcolor{blue}{\textbf{84.89}} & 90.82 & 79.24 & 65.62
        & 88.08 & 92.07 & 99.01 & 90.03& 81.87
        & 90.62 & 93.13 & 99.36 & 91.86& 84.94
        & 73.53 & 60.56 & 88.61 & 66.41& 49.72\\
        ChangeFormer~ \cite{cf} 
        & 77.08 & 79.37 & 89.87 & 78.21 & 64.72
        & 87.04 & 90.68 & 98.88 & 88.83& 79.90
        & 87.11 & 93.73 & 99.26 & 90.30& 82.32
        & 72.85 & 57.90 & 87.72 & 64.52& 47.63\\
        RS-Mamba~ \cite{rsmamba}
        & 71.04 & 81.89 & 89.47 & 76.08 & 61.40
        & 88.23 & 91.36 & 98.97 & 89.77& 81.44
        & 91.18 & 91.48 & 99.31 & 91.34& 84.05
        & \textcolor{red}{\textbf{81.93}} & 53.58 & 84.87 & 64.79& 47.92\\
        ChangeMamba~ \cite{chen2024changemamba} 
        & 76.54 & 82.43 & 90.62 & 79.38 & 65.80
        & 88.78 & 91.59 & 99.01 & 90.16& 82.09
        & \textcolor{blue}{\textbf{92.54}} & 92.28 & 99.39 & 92.41& 85.90
        & 61.80 & \textcolor{blue}{\textbf{70.60}} & \textcolor{blue}{\textbf{89.14}} & 65.91& 49.15\\
        CDMamba~ \cite{cdmamba}
        & \textcolor{red}{\textbf{78.68}} & 82.65 & 91.08 & \textcolor{blue}{\textbf{80.62}} & \textcolor{blue}{\textbf{67.53}}
        & 88.85 & 91.76 & 99.00 & 90.28& 82.28
        & \textcolor{red}{\textbf{92.94}} & 92.30 & 99.41 & 92.62& 86.26
        & 64.90 & 67.89 & \textcolor{red}{\textbf{89.92}} & 66.36& 49.66\\
        \midrule
        Ours
        & 76.03 & \textcolor{red}{\textbf{86.65}} & \textcolor{blue}{\textbf{91.59}} & \textcolor{red}{\textbf{80.99}} & \textcolor{red}{\textbf{68.06}}
        & \textcolor{blue}{\textbf{90.06}} & 90.98 & \textcolor{blue}{\textbf{99.01}} & \textcolor{red}{\textbf{90.52}} & \textcolor{red}{\textbf{82.68}}
        & 90.56 & \textcolor{red}{\textbf{96.02}} & \textcolor{blue}{\textbf{99.48}} & \textcolor{red}{\textbf{93.21}} & \textcolor{red}{\textbf{87.28}}
        & 73.12 & 61.95 & 89.00 & \textcolor{red}{\textbf{67.07}} & \textcolor{red}{\textbf{50.45}}\\
        \bottomrule
    \end{tabular}
    
    \vspace{-4pt}
    \label{tab1}
\end{table*}

\begin{figure*}
  \includegraphics[width=\textwidth]{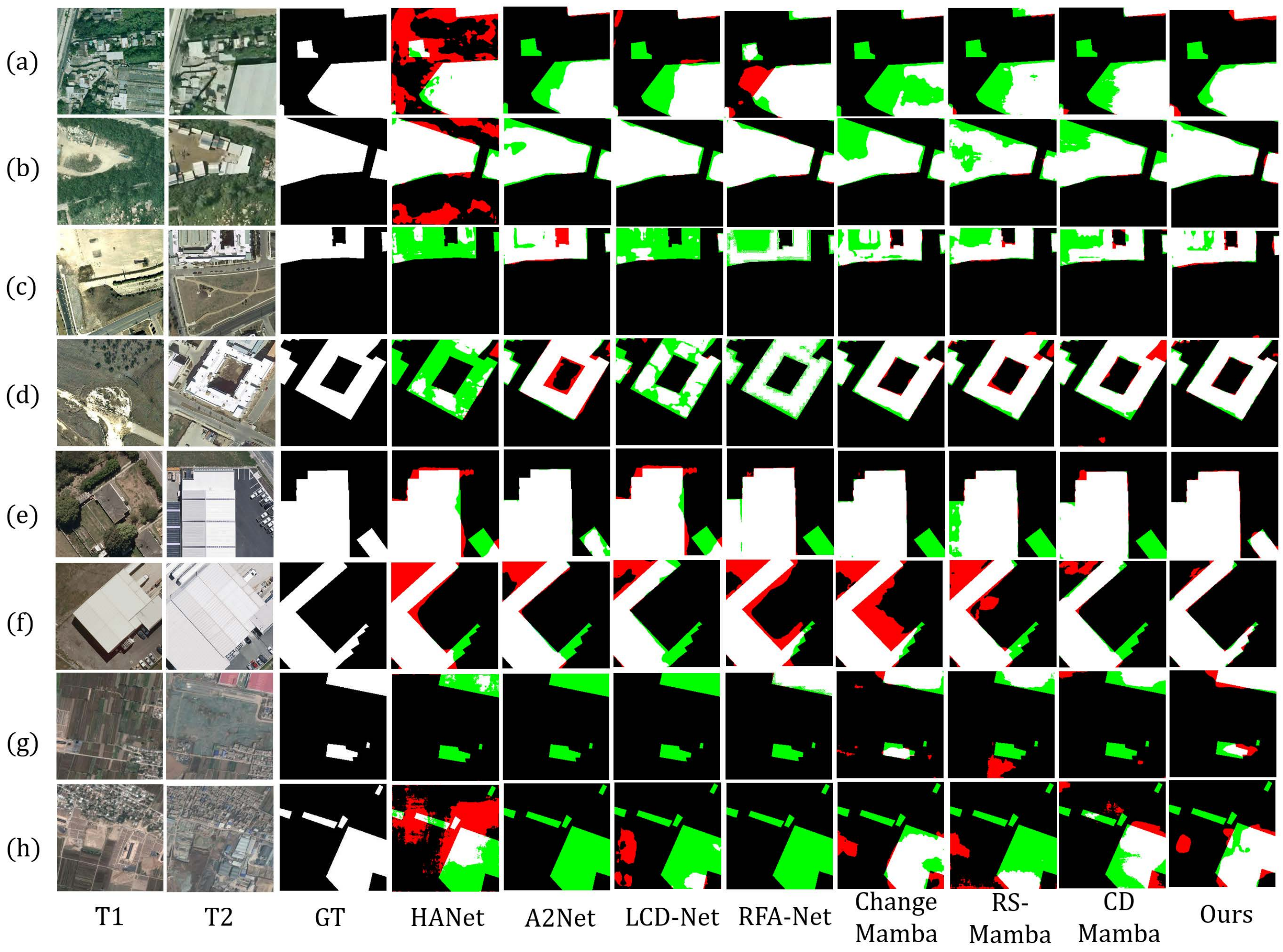}
  \caption{Qualitative results comparison across four datasets. White represents True Positives (TP), red represents False Positives (FP), green represents False Negatives (FN), and black represents True Negatives (TN). The leftmost three columns show bi-temporal images and ground truth.  All method names are labeled in the bottom row. Cases (a)-(b) are from SYSU-CD, (c)-(d) from LEVIR-CD, (e)-(f) from WHU-CD, and (g)-(h) from DSIFN-CD datasets.}
  \label{fig3}
\end{figure*}

\section{METHODOLOGY}

The overall architecture of the LDGNet is shown in \hyperref[fig2]{Fig. 2}. Unlike traditional Siamese encoder structures, our model employs a dual-branch encoder without weight sharing. One branch is dedicated to extracting multi-scale features from absolute difference images, while the other inherits the functionality of traditional Siamese encoders to extract features from original optical images. These two branches interact through the Difference Guiding Module (DGM), establishing a hierarchical guiding mechanism.
At the decoder stage, we introduce Visual State Space Model (VSSM) for the first time in lightweight optical change detection decoding. By embedding it within Difference-Aware Dynamic Fusion (DADF) modules, we utilize the absolute differences of extracted features to guide more targeted decoding. We integrate two fundamental feature modeling approaches (feature concatenation and feature difference) through a gating mechanism for dynamic fusion, and progressively upsample to obtain the final prediction map. 

\subsection{Encoding with Absolute Difference}
The encoder of LDGNet adopts dual MobileNetV3 \cite{mobilenetv3} backbones, leveraging their computational efficiency while enhancing their limited feature representation capacity through a difference guiding mechanism. Building upon conventional approaches, we incorporate the pixel-wise absolute difference between pre-event and post-event images – $Image^{diff}$ – as the model's extra input channel, which is defined as follow:
\begin{equation}
    Image^{diff}=\left | Image^{pre}-Image^{post} \right |
\end{equation}
Unlike existing approaches that compute feature differences after extraction, our design choice is motivated by the observation that compared with subtracting features extracted by the original image encoder, directly computing pixel-level differences from raw images can more comprehensively preserve semantic information and more explicitly represent the distribution of pixel variations. 

Also, we implement a dedicated encoder for processing difference images that operates independently without weight sharing with the Siamese encoders handling the original images. This architectural decision prevents potential feature contamination that could occur through parameter sharing between the difference and original image processing pathways. Also, it allows the difference encoder to develop unbiased representations of pixel-level variation patterns that can effectively guide and enhance the original encoders' sensitivity to meaningful changes.

We partition the complete MobileNetV3 into four hierarchical layers, where the encoder processing absolute difference images utilizes its first three layers, while the encoder handling original images employs all four layers. Specifically, the absolute difference image branch extracts multi-scale difference features $\left \{ F^{diff} \right \} _{j=1}^3$ across three stages. For each layer, the original image branch first extracts both pre-event features $\left \{ F_{ori}^{pre} \right \} _{j}$ and post-event features $\left \{ F_{ori}^{post} \right \} _{j}$, which are then modulated with the difference features $\left \{ F^{diff} \right \} _{j}$ via the DGM to produce the layer's final output—modulated features $\left \{ F_{mod}^{pre} \right \} _{j}$ and $\left \{ F_{mod}^{post} \right \} _{j}$. The network comprises four such encoder layers connected in a cascaded manner, where the output of the preceding layer serves as the input to the next, ensuring progressive propagation of differential information to guide hierarchical feature encoding. This process can be expressed as follows:

\begin{equation}
    \left \{ F_{ori} \right \} _{j}=MobileNetLayer{\left \{ F_{mod} \right \} _{j-1}}
\end{equation}

\begin{equation}
    \left \{ F_{mod} \right \} _{j}=DGM\left \{{\left \{ F_{ori} \right \} _{j}}\ ,\ { \left \{ F^{diff} \right \} _{j}}\right \}
\end{equation}
Please note that for the last layer's feature map, we no longer perform modulation on it, because there is no need to guide subsequent feature extraction.

DGM consists of two key components: the Difference Adapter (DA) module and the Spatial-Channel Attention (SCA) module. Their detailed architectures are illustrated in \hyperref[fig2]{Fig. 2(b)}.

DA employs depthwise separable dilated convolutions to expand the receptive field, effectively capturing both local details and global context with minimal computational overhead. It aligns difference features with original features for fusion, the residual connection \cite{resnet} ensures the stable propagation of the difference information. The output of the DA module $F_{DA}$ is fed to the SCA module.

In the SCA module, the attention mechanisms for both spatial \cite{spatial} and channel \cite{channel} dimensions are refined to meet the specific needs of difference guidance. In the spatial attention mechanism, we incorporate max-pooling operations to amplify the influence of differential regions, ultimately deriving the attention weights $A_{spatial}$ through a $3\times3$ convolutional layer and a sigmoid activation function.
The channel attention component $A_{channel}$ extracts global statistical information through global average pooling and then generates the channel attention vector via a two-layer fully connected network with nonlinear activation. SiLU \cite{silu} activation preserves negative-value information crucial for detecting change-absent regions (e.g., disappeared objects).
By fusing these two attention components, the final output $F_{SCA}$ of the SCA module is given by:

\begin{equation}
    F_{SCA}=\alpha \odot A_{spatial} \odot A_{channel}\odot F_{DA}   
\end{equation}
where $\alpha$ is a learnable parameter designed to adjust the degree to which the difference features influence the original features. $\odot$ denotes element-wise multiplication. 

Ultimately, the DGM fuses the original features with the difference features $F_{SCA}$ processed by the DA and SCA modules, with element-wise multiplication with residual connections.

The modulated features $\left \{ F_{mod} \right \} _{j}$ are retained and subsequently fed into the original image encoder at the next layer, thereby establishing a hierarchical difference-guided mechanism. This ensures that feature extraction at each layer is influenced by the difference features from the preceding layer, progressively refining the representation through multi-level guidance.

\subsection{Decoding with Difference Guidance}\label{sec3c}
Existing lightweight change detection decoders typically employ indiscriminate CNN-based modeling of feature maps, which not only forces the model to learn substantial irrelevant information and waste computational resources, but also remains constrained by limited receptive fields. Also, current approaches focus on two separate fundamental modeling directions - feature differencing and feature concatenation. Some methods attempt to sum the resultant maps obtained from these two directions, but few have explored their dynamic integration. Our Difference-aware Dynamic Fusion (DADF) module addresses these challenges by: (1) applying feature absolute difference-based weighting to enhance discriminative regions for more targeted modeling; (2) adopting the emerging VSSM architecture to achieve global relation modeling with linear computational complexity; and (3) generating difference attention maps to dynamically combine feature differencing and concatenation approaches, thereby emphasizing change regions while suppressing background noise.

Detailed architecture of DADF is shown in \hyperref[fig2]{Fig. 2(c)}. First, We computes the differences between the modulated pre-event and post-event features to generate refined difference features $\hat F^{diff}$  , with the specific computation as:

\begin{equation}
\hat F^{diff} = \left | F_{mod}^{pre} - F_{mod}^{post} \right |
\end{equation}
It should be noted that we utilize pixel-wise subtraction of the encoder-extracted features as the guidance signal during decoding. This approach emphasizes object-level changes rather than pixel-level variations. The feature-based difference maps provide semantic-level guidance for the decoding process, which facilitates the generation of more complete and accurate change prediction maps.

Subsequently, the network applies element-wise weighting to the original features using the difference features $\hat F^{diff}$ to obtain the weighted pre-event $\left \{F_{weighted}^{pre}\right \}_{j}$ and post-event features $\left \{F_{weighted}^{post}\right \}_{j}$. Also, we retain residual connections to facilitate the model's learning of difference relationships between pre-event and post-event features.

To further model spatio-temporal information, these two sets of weighted features, $\left \{F_{weighted}^{pre}\right \}_{j}$ and $\left \{F_{weighted}^{post}\right \}_{j}$ are concatenated along the channel dimension. They are first reduced in dimensionality using a $1\times1$ convolution and then fed into the Visual State Space (VSS) block, with the intermediate feature representation:

\begin{align}
P'=VSSBlock(Conv_{1\times1}( \left \{ F_{weighted}^{pre} \right \} _{j} \operatornamewithlimits{\textcircled{c}}\left \{ F_{weighted}^{post} \right \} _{j}  ))
\end{align}

The detailed structure of the VSS Block \cite{liu2024vmamba} is illustrated in \hyperref[fig2]{Fig. 2(c)}. Its core component is the SS2D \cite{liu2024vmamba} module, which performs directional scanning across the image, ensuring that each pixel captures comprehensive global semantic information with linear computational complexity. Details of SS2D can be found in \cite{liu2024vmamba}.

DADF introduces a difference attention mechanism. For each layer's difference feature $\left \{\hat F^{diff}\right \}_j$, a convolution followed by a Sigmoid activation is used to generate the difference attention map $(DiffAttn)_{j}$. Subsequently, we employ this attention mechanism as weighting to integrate the intermediate feature $P'$ with the difference features $\hat F^{diff}$, resulting in the fused feature map $P$ at the current level. The corresponding formulation is as follows:

\begin{align}
P=DiffAttn\odot P' +(1-DiffAttn)\odot \hat F^{diff} + P'
\end{align}
when the DiffAttn is high, the region is considered more significant in change detection, prompting the model to rely more on P' for change identification, preserves region-specific features highly correlated with changes. Conversely, when the DiffAttn is low, indicating minimal or no changes in the region, the model primarily depends on the difference feature for refinement and captures latent subtle variations. Residual connections prevent excessive disturbance. This step effectively integrates the two predominant modeling paradigms in change detection, feature difference and feature concatenation, by dynamically adjusting their weighting coefficients to achieve complementary advantages. The adaptive fusion mechanism simultaneously preserves salient change-related information while suppressing background interference and noise artifacts.

After performing dual fusion on each extracted feature layer, we upsample and sum the final feature maps $P$ from each layer, then pass them through a $1\times1$ convolution layer to obtain the ultimate feature map.

\section{Experiment}
\subsection{Experimental Settings}
\textbf{Datasets:} This study employs four benchmark datasets to comprehensively validate the model's multi-scenario adaptability. The SYSU-CD \cite{sysu} dataset is specifically designed to evaluate comprehensive performance in complex multi-category change detection. Both LEVIR-CD \cite{LEVIR} and WHU-CD \cite{whu} datasets focus on building-level fine-grained change identification. The DSIFN-CD \cite{dsifn} dataset addresses the requirements for large-scale urban change analysis. All datasets were uniformly preprocessed into standardized input of 256×256 pixels.

\textbf{Implementation Details:} Experiments were conducted on an NVIDIA RTX 3090 GPU (24GB VRAM) with a batch size of 16. The AdamW optimizer was employed with an initial learning rate of 0.0001 and weight decay coefficient of 0.0005. We use the compound loss function combining Cross-Entropy Loss $\mathcal{L}_{ce}$ and Lovász-Softmax Loss \cite{berman2018lovasz} $\mathcal{L}_{lov}$, it is formally defined as:
\begin{align}
\mathcal{L}_{final}=\mathcal{L}_{ce}+\mathcal{L}_{lov}
\end{align}

\textbf{Compared Methods:} Our experiment systematically selects representative lightweight and high-parameter models, encompassing three predominant architectural paradigms: convolutional neural network (CNN)-based architectures, Transformer-based architectures, and the emerging state space model (SSM)-based architectures (Mamba series). We choose FC-EF \cite{daudt2018fully}, FC-Siam-Diff \cite{daudt2018fully}, FC-Siam-Conc \cite{daudt2018fully}, SNUNet \cite{fang2021snunet}, A2Net \cite{a2}, BIT \cite{BIT}, LCD-Net \cite{lcd}, RFANet \cite{rfa} as lightweight methods, and IFNet \cite{ifnet}, DTCDSCN \cite{dtcdscn}, HANet \cite{han2023hanet}, MSPSNet \cite{MSPS}, STANet \cite{sta}, ChangeFormer \cite{cf}, RS-Mamba \cite{rsmamba}, ChangeMamba \cite{chen2024changemamba}, CDMamba \cite{cdmamba} as heavyweight methods.

\textbf{Evaluation\ Metrics:} We evaluate the model's performance using five metrics: recall rate (\textbf{Rec}), precision rate (\textbf{Pre}), overall accuracy (\textbf{OA}), F1 score (\textbf{F1}), and intersection over union (\textbf{IoU}) \cite{zhibiao1,zhibiao2,zhibiao3}. Here we make \textbf{F1} and \textbf{IoU} main evaluation metrics.

\subsection{Comparisons with SOTA}

\textbf{Quantitative Comparison.}
\hyperref[tab1]{Table 1} presents quantitative comparison results. Among all comparative methods, our approach achieves superior performance across multiple metrics. Notably, it attains the best performance in both F1 and IoU on all datasets. It is interesting that on the cross-domain DSIFN-CD dataset (with training and test sets from distinct cities), existing methods exhibit prevalent imbalance between omission and commission errors , e.g., RS-Mamba shows 81.93\% recall but only 53.58\% precision, LCD-Net shows 77.84\% precision and 54.14\% recall. Our method achieves better-balanced optimization. For the multi-category SYSU-CD dataset, our method demonstrates comprehensive leadership across three metrics, validating its adaptability to complex scenarios. Finally, on building-specific datasets like LEVIR-CD and WHU-CD, our method also demonstrates competitive performance, achieving top-2 rankings in four out of five evaluation metrics.

\textbf{Qualitative Comparation.}
As demonstrated in \hyperref[fig3]{Fig. 3}, through visual comparisons, our method exhibits significant advantages in complex scenarios:  In \hyperref[fig3]{Case (e)}, our method achieves complete detection of buildings with camouflage colors (lower-right region), while 
 almost all comparative methods show varying degrees of omission errors. \hyperref[fig3]{Case (f}) validates the method's robustness against irrelevant interference. Our method effectively suppresses pseudo-change interference caused by surface color variations. In \hyperref[fig3]{Case (g)} and \hyperref[fig3]{(h)}, the three compared lightweight detection methods almost fail to effectively extract meaningful features in large-scale urban changes, while our method successfully overcomes this limitation.

\begin{table}[t!]
\centering
\caption{Computation Cost Comparison of Different Methods. Params denotes the model's parameter count, FLOPs represents Floating Point Operations, Mem represents the peak GPU memory consumption during inference. The lowest computational resources are highlighted in \textcolor{red}{\textbf{red}}. All input tensors are of size (1, 3, 256, 256).}
\setlength{\tabcolsep}{4pt}
\begin{tabular}{c|c|ccc}
\toprule
\textbf{Type} & \textbf{Method} & \multicolumn{1}{c}{\textbf{Params}} & \multicolumn{1}{c}{\textbf{FLOPs}} & \multicolumn{1}{c}{\textbf{Mem}} \\
 & & \multicolumn{1}{c}{\textbf{(M)}} & \multicolumn{1}{c}{\textbf{(G)}} & \multicolumn{1}{c}{\textbf{(MB)}} \\
\midrule
\multirow{8}{*}{Lightweight} 
 & FC-EF & \textcolor{red}{\textbf{1.34}} & 3.58 &1033 \\
 & FC-Siam-Diff & 1.35 & 4.73 &1108\\
 & FC-Siam-Conc & 1.54 & 5.33 &1648\\
 & SNUNet & 1.35 & 4.72 &552 \\
 & BIT & 11.89 & 8.75 &2128\\
 & A2Net & 3.78 & 6.02 &906\\
 & LCD-Net & 4.45 & 2.56 &7238\\
 & RFANet & 2.86 & 3.16 &3308 \\
\midrule
\multirow{8}{*}{Heavyweight}
 & IFNet & 50.71 & 41.18 &8508\\
 & DTCDSCN & 41.07 & 14.42 &1105 \\
 & HANet & 2.61 & 17.67 &3459 \\
 & MSPSNet & 1.79 & 15.04 &3458 \\
 & ChangeFormer & 41.03 & 202.86 &8543 \\
 & RS-Mamba & 42.30 & 18.36 &2621 \\
 & ChangeMamba & 32.52 & 20.50 &2716 \\
 & CDMamba & 11.90 & 49.68 &2149 \\
\midrule
Ours & LDGNet & 3.43 & \textcolor{red}{\textbf{1.12}} &\textcolor{red}{\textbf{513}} \\
\bottomrule
\end{tabular}
\label{tab2}
\end{table}

\textbf{ Computational Overhead Analysis.}
 As evidenced by the results in \hyperref[fig:comparison]{Fig. 1}, our method achieves superior F1 scores while utilizing significantly fewer parameters (3.43M) and FLOPs (1.12G). Notably, our method demonstrates the lowest computational complexity in both FLOPs and peak GPU memory consumption among all benchmarked methods. Detailed computation cost is shown in \hyperref[tab2]{Table 2}.
Peak memory consumption does not linearly correlate with parameter count. Certain lightweight methods exhibit unexpectedly high memory demands (e.g., LCD-Net requires 7238MB), whereas our approach achieves optimal memory efficiency (513MB), making it particularly suitable for real-world deployment scenarios.

\begin{figure}
  \includegraphics[width=0.4\textwidth]{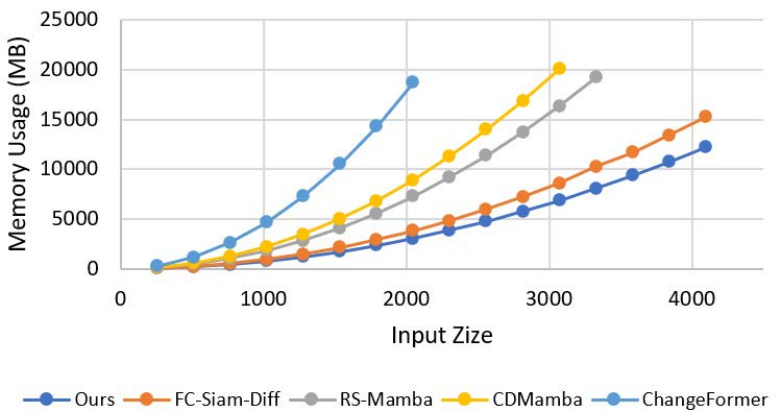}
  \caption{Comparison of GPU memory consumption across different methods as a function of input image size. }

  \label{fig4}
\end{figure}

 In practical applications, directly processing large-scale remote sensing images is crucial. As image size increases, our method maintains manageable memory consumption—comparable to traditional lightweight convolutional approaches, as shown in \hyperref[fig4]{Fig. 4}. In contrast, some Transformer-based and pure Mamba architectures face out-of-memory (OOM) risks.

\subsection{Ablation Study}
To validate the effectiveness of our proposed DGM and DADF modules, we designed ablation experiments to evaluate the detection performance and robustness against interference on the WHU dataset. We first established a baseline by removing both DGM and DADF while retaining only single MobileNetV3 backbone and VSSM. Results are shown in \hyperref[tab3]{Table 3}. The baseline achieved an exceptionally high Recall of 93.69\%. However, its Precision remained at only 90.04\%, indicating significant false positive predictions that degraded overall performance. Meanwhile, the performance of the baseline model deteriorates significantly with the introduction of Gaussian noise and blurring. Using only lightweight backbone network and VSSM for change detection will inevitably lead to a certain degree of performance degradation.

\begin{table}[!htbp]
\centering
\caption{Ablation study of DGM and DADF modules on WHU dataset, with best metrics highlighted in \textcolor{red}{red}.}
\label{tab3}
\begin{tabular}{cc|ccccc}
\toprule
\multicolumn{2}{c|}{Modules} & \multicolumn{5}{c}{Metrics (\%)} \\ 
\cmidrule{1-7} 
DGM & DADF & Rec & Pre & OA & F1 & IoU \\
\midrule
\ding{55} & \ding{55} & \textcolor{red}{\textbf{93.69}} & 90.04 & 99.33 & 91.83 & 84.90 \\
\ding{51} & \ding{55} & 91.70 & 93.32 & 99.41 & 92.51 & 86.06 \\
\ding{55} & \ding{51} & 92.21 & 93.26 & 99.43 & 92.73 & 86.45 \\
\ding{51} & \ding{51} & 90.56 & \textcolor{red}{\textbf{96.02}} & \textcolor{red}{\textbf{99.48}} & \textcolor{red}{\textbf{93.21}} & \textcolor{red}{\textbf{87.28}} \\
\bottomrule
\end{tabular}
\vspace{0.2cm}

\end{table}

\begin{figure}
  \includegraphics[width=0.45\textwidth]{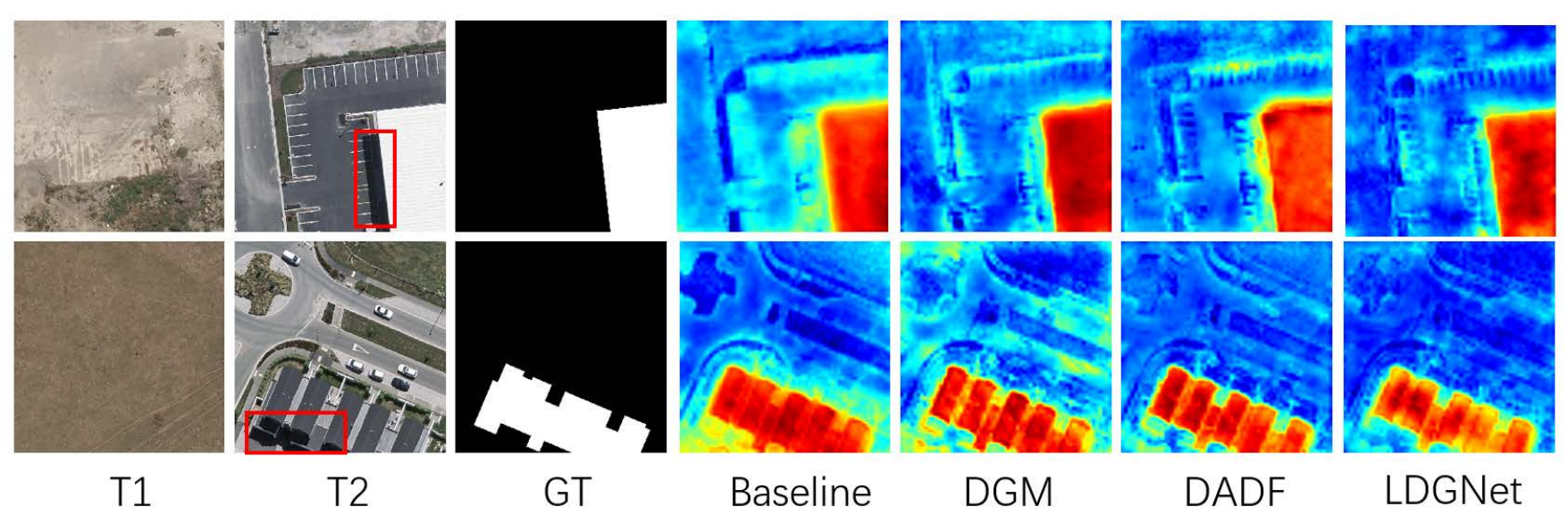}
  \caption{Heatmap comparison. Response intensity in shadow areas (marked by red boxes in T2) is weaken. }
  \label{fig5}
\end{figure}

\begin{figure}[htbp]
  \centering
  \begin{subfigure}[t]{0.48\columnwidth} 
    \includegraphics[width=\linewidth,height=3cm]{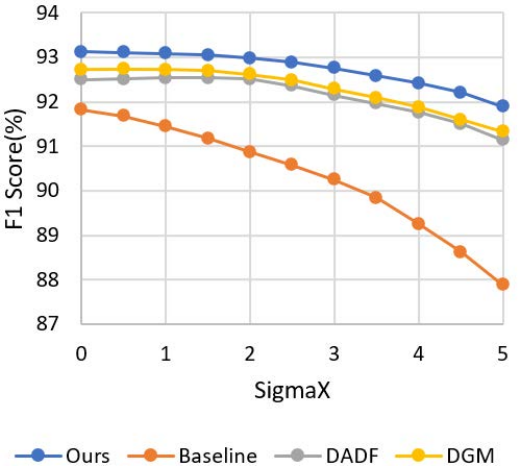}
    \caption{Gaussian noise }
    \label{fig6}
  \end{subfigure}
  \hfill
  \begin{subfigure}[t]{0.48\columnwidth}
    \includegraphics[width=\linewidth,height=3cm]{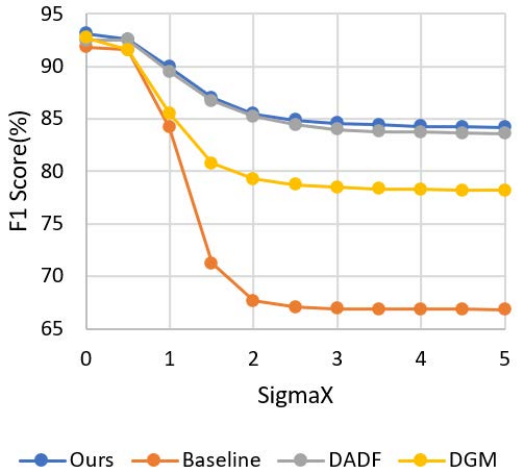}
    \caption{Gaussian blur}
    \label{fig:b}
  \end{subfigure}
  \caption{Interference intensity and inference results. X-axis represents the standard deviation of the Gaussian function. Gaussian blur kernel size is fixed at 3×3. }
  \label{fig:parallel}
\end{figure}

\textbf{Effect of DGM:} We introduced a hierarchical guidance mechanism based on absolute difference images and original optical images, ultimately achieving a better balance between Recall and Precision, thereby improving the F1 score and IoU by 0.9\% and 1.55\%, as shown in \hyperref[tab3]{Table 3}.
Visualization results in \hyperref[fig5]{Fig. 5} demonstrate that the model exhibits reduced responses to shadowed regions. Additionally, \hyperref[fig6]{Fig 6} shows its enhanced robustness against Gaussian noise and achieves moderate improvements in handling Gaussian blur.

\textbf{Effect of DADF:} More targeted difference-guiding modeling and dynamic fusion of feature differences with feature concatenation can improve accuracy to some extent. Visualization results in \hyperref[fig5]{Fig 5} show that DADF further suppresses shadow interference while producing sharper building boundaries. Although its performance improvement is slightly inferior to DGM, DADF demonstrates strong resilience against both high-frequency and low-frequency interference. When the standard deviation of Gaussian noise and Gaussian blur reaches 5, the baseline suffers a relative F1-score degradation of 4.2\% and 27.2\%, respectively. In contrast, DADF effectively confines the performance drop within 1.4\% and 15.6\%, demonstrating significantly better robustness.

\section{Conclusion}
This study proposes LDGNet (Lightweight Difference Guiding Network). This architecture aims to address the challenge of limited feature representation capacity in lightweight networks through adaptive guiding mechanism integrating absolute difference images and original images. Also, we achieve more targeted global modeling enabled by feature differences guidance and dynamic fusion. The innovative design incorporates a Difference Guiding Module (DGM) and Difference-Aware Dynamic Fusion (DADF), establishing full-process difference guidance from encoding to decoding. Experimental results demonstrate its state-of-the-art detection accuracy under computational constraints of 1.6G FLOPs and 513MB peak memory consumption. Meanwhile, the interaction between difference features and original features enhances the robustness of feature extraction and modeling, leading to superior suppression of noise and background interference. However, the dual-MobileNet encoder structure inevitably introduces additional parameters compared to traditional Siamese encoders. In future work, we plan to explore more parameter-efficient approaches. This work aims to strike an optimal balance between model complexity and detection performance, ultimately facilitating broader deployment of deep learning-based change detection in real-world scenarios through enhanced computational efficiency.

\clearpage
\bibliographystyle{ACM-Reference-Format}
%\bibliography{sample-base}

\appendix

\end{document}